\documentclass[sigconf]{acmart}

\usepackage{graphicx}
\usepackage{epstopdf}
\usepackage{dblfloatfix}
\usepackage{booktabs}
\usepackage{enumitem}
\usepackage{multirow}
\usepackage{url}
\usepackage{breakurl}
\usepackage{subcaption}
\usepackage{color}
\usepackage[font=footnotesize]{caption} 

\usepackage[ruled, linesnumbered]{algorithm2e}
\usepackage{bm}

\begin{document}
%
\title{Sensor-Aware Classifiers for Energy-Efficient Time Series Applications on IoT Devices}
\author{Dina Hussein}
\email{dina.hussein@wsu.edu}
\affiliation{
\institution{Washington State University}
\city{Pullman}
  \state{WA}
   \country{USA}}
   
\author{Lubah Nelson}
\email{lubah.nelson@wsu.edu}
\affiliation{
\institution{Washington State University}
\city{Pullman}
  \state{WA}
   \country{USA}}
   
\author{Ganapati Bhat}
\email{ganapati.bhat@wsu.edu}
\affiliation{
\institution{Washington State University}
\city{Pullman}
  \state{WA}
   \country{USA}}

\begin{abstract}
Time-series data processing is an important component of many real-world applications, such as health monitoring, environmental monitoring, and digital agriculture.
These applications collect distinct windows of sensor data (e.g., few seconds) and process them to assess the environment.
Machine learning~(ML) models are being employed in time-series applications due to their generalization abilities for classification.
State-of-the-art time-series applications wait for entire sensor data window to become available before processing the data using ML algorithms, resulting in high sensor energy consumption. 
However, not all situations require processing full sensor window to make accurate inference.
For instance, in activity recognition, sitting and standing activities can be inferred with partial windows.
Using this insight, we propose to employ early exit classifiers
with partial sensor windows to minimize energy consumption while maintaining accuracy.
Specifically, we first utilize multiple early exits with successively increasing amount of data as they become available in a window. If early exits provide inference with high confidence, we return the label and enter low power mode for sensors.
The proposed approach has potential to enable significant energy savings in time series applications. We utilize neural networks and random forest classifiers to evaluate our approach.
Our evaluations with six datasets show that the proposed approach enables up to 50-60\% energy savings on average without any impact on accuracy. The energy savings can enable time-series applications in remote locations with limited energy availability.
\end{abstract}
\maketitle



%

\section{Introduction}
Time-series data processing is an integral component of several critical applications such as health monitoring, environmental sensing, digital agriculture, and wildfire monitoring~\cite{lara2012survey,espay2016technology,shoaib2014fusion,vasisht2017farmbeats}.
The applications utilize one or more sensors to monitor environmental parameters or user's health~\cite{shoaib2014fusion,estrin2010open,ozanne2018wearables,
daneault2018could}.
For example, wildfire monitoring applications utilize humidity and temperature sensors~\cite{kaur2019energy,brito2020wireless}, while health monitoring applications may use motion, heart rate, or electrodermal activity sensors~\cite{espay2016technology,hughes2023wearable,moore2008ambulatory}. 
The sensor data are then processed in distinct windows that range from milliseconds to few seconds to obtain desired outcomes for the application~\cite{lovric2014algoritmic,wang2016comparative}.
Processing using windows allows the application to quickly react to changes in conditions and maintain temporal resolution~\cite{lovric2014algoritmic,lemire2007better}.

Time-series applications typically accumulate sensor data for the window duration for processing.
For instance, the activity recognition approach in~\cite{shoaib2014fusion} uses 10~s windows for processing. This means that sensor data are first collected for 10 seconds before identifying the activities. Once data are accumulated they are passed on to feature generation and application processing blocks. 
Machine learning~(ML) algorithms are becoming popular in time series applications. 
Some popular ML models used in time-series applications include neural networks, random forests, and decision trees~\cite{ahmed2010empirical}.

ML models typically collect data for the appropriate window before using them in classification~\cite{teerapittayanon2016branchynet}.
One of the key requirements of traditional ML models is that data for entire window must be available for processing.
This is because model architectures are structured and trained with the assumption that data from all sensors are available for each window. 
Consequently, the sensors must be turned on for entire window duration, which increases the system energy consumption. Indeed, sensor energy dominates overall energy consumption in time-series Internet of Things (IoT) applications~\cite{bhat2018online,mochizuki2021power,du2020energy,
martinez2015power}. 
The high energy consumption is a significant challenge for low-power IoT devices that do not have large battery capacities. Even in devices with large batteries, minimizing sensor energy consumption will improve the overall quality of service.

Prior research has proposed several methods to minimize energy consumption of ML models and time series applications~\cite{kansal2007power,mochizuki2021power,du2020energy,agbehadji2020intelligent}. These methods include duty cycling the sensors, reducing the sampling rate, or utilizing fewer sensors.
However, each of these methods may lead to reduction in accuracy since all possible sensor data are not being used. Furthermore, it may require design of multiple ML models to handle multiple duty cycling and sampling rate options, thus adding to memory and context switching overhead at runtime.
Recent research has also proposed early exit neural networks that include multiple exit points for a given ML model~\cite{teerapittayanon2016branchynet,ju2021dynamic,li2023predictive}. The key idea in early exit neural networks is to utilize earlier exits if the model is confident of its prediction. This is useful to save energy when classifying easier inputs since all layers are not executed.
However, early exit neural networks still require the complete sensor data and do not reduce sensing energy. Therefore, there is a strong need to investigate approaches that minimize sensor energy consumption in time series applications.

This paper presents a novel sensor-aware early exit approach, referred to as SEE, for time-series applications.
The key idea behind the proposed approach is to iteratively feed portions of sensor data into the ML models and utilize early exits to assess prediction confidence.
If the model is confident with partial data at the early exit, SEE returns the prediction and turns off sensors until the next window, leading to significant energy savings.
On the other hand, if desired confidence is not achieved at the early exit, additional data are fed into the model for processing. For instance, consider a neural network with five layers and early exits at layers one and two, respectively. SEE first provides 20\% of data in a window and checks the confidence at exit one. If the confidence is below a threshold, SEE provides additional 20\% of data for layers two and beyond.
Using early exits with partial data has the potential to enable significant energy savings without compromising accuracy.

The proposed approach is based on two key insights as follows: \textit{first,} not all labels or classes in an application require data from the entire window for accurate inference. For instance, sitting and standing in activity recognition can be recognized with partial data since they present constant motion values, while complex activities require full data sequence.
\textit{Second,} we can design the ML models such that they have multiple input and exit layers to perform processing with partial data. 
Leveraging these insights enables SEE to provide energy savings by turning off sensors once required confidence levels for classification are met.

We validate the proposed approach with six diverse activity and health datasets when using convolutional neural networks and random forest as the ML models. We choose activity and health as our driver applications since health monitoring using low-power wearable devices has the potential to transform healthcare~\cite{espay2016technology,maetzler2016clinical}. These devices typically operate under energy constraints, thus motivating a need for efficient processing.
We train the proposed SEE neural networks for each dataset by varying the exit locations and thresholds for confidence. 
Our evaluations with all datasets show that the proposed approach achieves accuracy comparable to the default accuracy with no early exits while reducing sensing energy by up to 50-60\% on average.
The overhead of SEE classifiers is also negligible when compared to potential of sensor energy savings.
Evaluation with random forest classifiers show similar results, thus highlighting the generalizability of the proposed approach.
In summary, this paper makes the following contributions:
\begin{itemize}
    \item Novel sensor-aware early exit architectures that take partial data inputs in each window and exit early when inference can be made with high confidence with partial window data,
    \item Late input blocks for SEE neural networks that combine information from previously used data and new sensor readings to ensure that information inputs given at the beginning of a window are not lost,
    \item Algorithms to train the sensor-aware early exit models to achieve high accuracy and energy efficiency,
    \item Experimental evaluations with six diverse datasets to show up to 50-60\% energy savings on average with the proposed approach while not reducing accuracy.
\end{itemize}

The rest of the paper is organized as follows: Section~\ref{sec:related_work} reviews the related work, while Section~\ref{sec:overview} provides an overview of the proposed approach.
We discuss the proposed SEE approach in Section~\ref{sec:il_method} and provide the experimental results in Section~\ref{sec:experiments}. Finally, section~\ref{sec:conclusion} concludes the paper with future research directions.

\section{Related Work} \label{sec:related_work}

Optimizing energy consumption in time-series applications and ML models is pivotal, especially with their growing integration into low power wearable devices \cite{jeon2023harvnet}. 
Balancing energy efficiency with accuracy presents a significant challenge, as strategies that reduce power and energy consumption often compromise data precision~\cite{li2023predictive}. This trade-off between accuracy and energy consumption has led to various approaches aimed at minimizing sensor operation and data processing energy demands without sacrificing the integrity of the results \cite{teerapittayanon2016branchynet}. 

Recently proposed methods to reduce ML model energy consumption, such as early exit neural networks, reduce computational energy by halting processing when predictions reach sufficient confidence levels~\cite{teerapittayanon2016branchynet,ju2021dynamic,jeon2023harvnet,li2023predictive}. For example, HarvNet, designed for energy-harvesting IoT devices, employs a neural architecture search (HarvNAS) to create energy-efficient multi-exit architectures and a dynamic inference policy (HarvSched) for varying energy conditions~\cite{jeon2023harvnet}. However, despite their computational energy efficiencies, early exit models like HarvNet heavily rely on initial sensor data collection~\cite{jeon2023harvnet}. This means that while computational energy is optimized, significant energy used for collecting and preparing sensor data in time-series applications is not addressed. As such, these networks reduce processing time and energy but fail to decrease the energy costs related to sensing and data acquisition.

Duty cycling and sampling rate adjustments are popular techniques used to reduce sensor energy consumption \cite{lee2019intermittent,kim2017mobility}. Duty cycling activates sensors only during specific intervals to conserve energy, but this can miss crucial data during downtime. Moreover, it can affect application accuracy since all data are not used for inference~\cite{kim2017mobility}. On the other hand, adjusting sampling rates reduces data frequency to save power, which can affect data precision and accuracy.
Intermittent learning enhances this approach in energy-harvesting systems, significantly boosting energy efficiency by up to 100\% and reducing training data requirements by 50\%. However, it requires meticulous management of data and algorithms to ensure operational effectiveness under stringent energy constraints~\cite{lee2019intermittent}. Moreover, intermittent learning does not address the need to obtain full sensor data in a segment to perform processing.
Therefore, there is a strong need to investigate approaches that minimize sensor energy consumption while maintaining accuracy.

The proposed sensor-aware early exit~(SEE) framework addresses these challenges by optimizing sensor energy use, extending the early exit concept to sensor data processing. 
Unlike earlier models that primarily reduce computational energy, SEE also minimizes sensor energy demands without compromising data integrity. It enables the system to make accurate predictions with partial data segments, thereby reducing unnecessary sensor operation. This capability significantly enhances the energy efficiency of IoT and mobile devices for time-series applications. 
Our experimental results show that the proposed approach indeed results in significantly lower sensor energy consumption while maintaining accuracy.

\section{Overview and Preliminaries}\label{sec:overview}
This section provides an overview of data processing in time-series applications, followed by a brief description of the proposed approach. 
We also provide motivational examples and key insights used in the proposed sensor-aware early exit models.

\subsection{Time Series Application Overview}
Time series data processing arises in several applications including health monitoring, wide area sensing, and environmental monitoring. We focus on healthcare applications involving multiple sensors that monitor physiological parameters. These sensors are integral for tracking health conditions such as movement disorders, vital sign fluctuations, and rehabilitation progress ~\cite{espay2016technology, hughes2023wearable, moore2008ambulatory}. At the same time, we note that the proposed approach is applicable to other time series applications as well.

One of the key differences in time series compared to image-based applications is that data arrive one sample at a time and have to be accumulated for processing. In general, time series applications go through the following steps for processing.


\vspace{1mm}
\noindent\textbf{Sensor Data Collection and Segmentation:} The first step in time-series applications is to obtain data from relevant sensors, such as motion, electrocardiogram, or heart rate.
This data is then divided into segments of fixed or variable lengths for further processing. For instance, prior activity monitoring applications use segment lengths ranging from 1~s to 10~s.
Data segmentation is important since it is not possible to infer application labels based on individual samples. Moreover, frequent processing with each sample leads to higher overhead. 
Additionally, segment length must be short enough to capture dynamic variations in the application. For instance, one-hour segments are not optimal for fitness tracking due to dynamic change in user activities.
In most applications, segments typically range from a few seconds to capture the required information effectively.
\vspace{1mm}
\noindent\textbf{Classification with ML Models:} The second major step in time-series applications is to use ML models for processing. Without loss of generality, we assume that the applications involve classification tasks.
As such, the ML models are used to predict the class label as a function of observed time-series data.
For instance, activity monitoring applications may predict if the user is walking or sitting.
Classification results are then provided to the users for analysis.

\subsection{Energy Challenges with Time Series Applications}
Commonly used ML models and application processing flows for time-series applications use entire segment of data for classification. More specifically, the ML models first wait for complete data accumulation in a segment before predicting the class labels.
The sensors are turned on for the entire segment duration, leading to high sensing energy.
Indeed, prior studies have shown that majority of energy consumption in IoT devices is from the sensors. 
The high energy consumption is a challenge for small form-factor IoT devices with small batteries. Therefore, there is a need for approaches that minimize sensor energy consumption while maintaining accuracy.


\begin{table}[t]
\caption{Classes Accuracy for Shoaib dataset with different input data lengths}
\label{tab:motivation}
\begin{tabular}{@{}lrrr@{}} \toprule
                         & \multicolumn{3}{c}{Input data length}                                      \\ 
\multirow{2}{*}{Classes} & \multicolumn{1}{c}{30\%} & \multicolumn{1}{c}{40\%} & \multicolumn{1}{c}{50\%} \\ \cmidrule(l){2-4} 
                         & \multicolumn{3}{c}{Accuracy (\%)}                                              \\ \midrule
Biking                   & 3                        & 59                       & 90                       \\
Down stairs              & 29                       & 35                       & 47                       \\
Jogging                  & 58                       & 51                       & 62                       \\ 
Sitting                  & 0                        & 99                       & 98                       \\
Standing                 & 72                       & 98                       & 98                       \\
Up stairs                & 19                       & 49                       & 78                       \\
Walk                     & 13                       & 2                        & 2   \\      \bottomrule              
\end{tabular}
\end{table}

We hypothesize that sensor data for the entire window are not required for all classes within an application. That is, a subset of labels can be classified with high accuracy without using all data in a given window. For instance, it is easier to classify sitting activities compared to walking activities.
We also perform an experiment to validate our hypothesis. Specifically, we first train a classifier that uses entire data segments to classify activities in the Shoaib dataset~\cite{shoaib2014fusion}. Then, we test the classifier with varying lengths of partial segments.
Note that the classifier is trained with complete segments and we test by substituting the last observed value for the partial segments.
Table~\ref{tab:motivation} illustrates the classification accuracy across different activities and segment lengths.
Notably, the sitting and standing activities maintain high accuracy even when only 40\% of the data is used. These results demonstrate that significant sensor energy savings can be achieved by employing partial segments for activities that are easier to classify.

\noindent\textbf{Key Insights:} We can obtain the following \textit{key insights} from the accuracy analysis:
\begin{enumerate}
    \item A subset of classes in time-series applications can be classified with partial segments without compromising accuracy.
    \item ML models can be trained to account for partial data segments and exit early whenever it is confident in classifying the data.
    \item  We can turn off sensors early once enough data are available for these labels and the ML model is able to classify them accurately.
\end{enumerate}

The proposed approach leverages these key insights to obtain sensor-aware early exit ML models to enable significant energy savings, as described in the following sections.

\begin{figure}[b]
    \centering
    \includegraphics[width=1\linewidth]{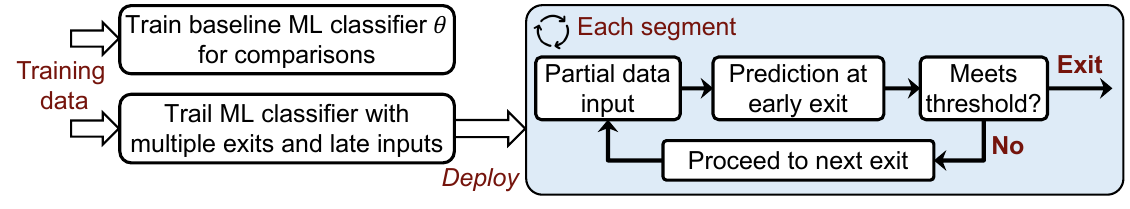}
    \vspace{-3mm}
    \caption{Overview of the SEE approach.}
    \label{fig:see_overview}
\end{figure}

\subsection{Overview of SEE Approach}
Figure~\ref{fig:see_overview} provides a high-level overview of the proposed approach. Initially, a base model processes entire  data segments to establish the highest possible accuracy. Then, we introduce sensor-aware models that analyze partial segments during initial processing. The model generates a classification based on the available data and assesses its confidence in the prediction. 
If the confidence level meets or exceeds a predetermined threshold, the predicted label is presented to the user, and the sensors are turned off to conserve energy. If confidence is insufficient, additional data are gathered for further analysis through early exits.
Late input modules are also incorporated to integrate more sensor data or contextual information obtained at later stages, ensuring a comprehensive analysis that leverages all relevant information.
This process continues until the terminal exit, where all available segment data are used. Sensor-aware early exit processing is repeated for each new data segment, optimizing energy savings while maintaining classification accuracy.

Key hyperparameters in the proposed approach include the number and placement of early exits, amount of partial segment data in each, and confidence thresholds. These hyperparameters must be carefully selected to balance energy savings and accuracy.
We utilize a design space exploration with the proposed models to determine the optimal parameters. Overall, the proposed approach leverages partial data segments and early exits in ML models to minimize sensor energy consumption.

\begin{figure}[b]
    \centering
    \includegraphics[width=1\linewidth]{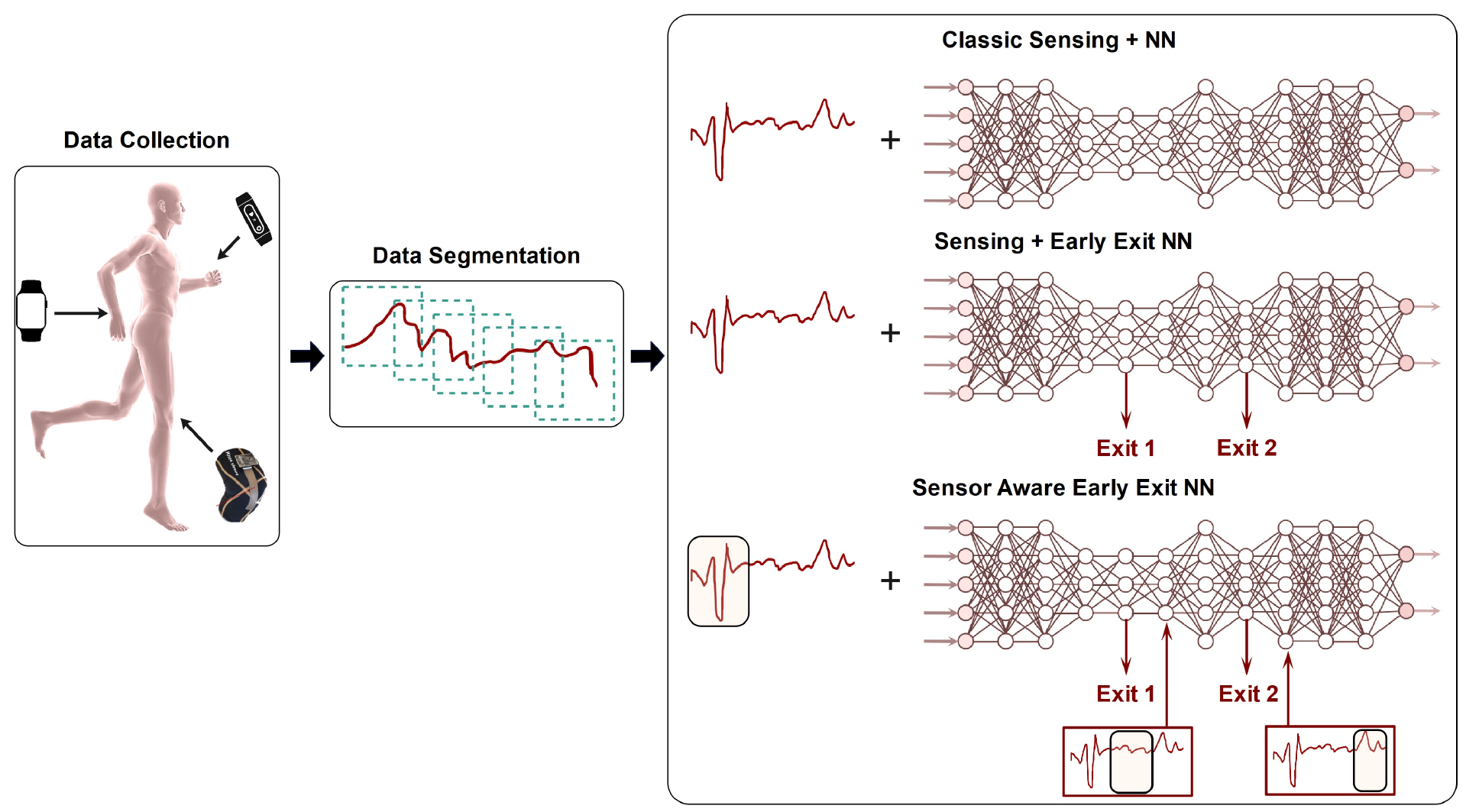}
    \caption{Overview of the proposed sensor aware neural networks}
    \label{fig:overview}
\end{figure}
\section{Sensor Aware Early Exit Framework}\label{sec:il_method}
This section explores the proposed sensor-aware early-exit models applied to both convolutional neural networks (CNNs) and random forest classifiers. 
We choose these two model types due to their widespread usage in time-series applications~\cite{hatami2018classification,tyralis2017variable}. 
They also offer complementary strengths in their performance. CNNs excel in extracting and processing spatial features from sensor data, while random forests provide robust performance across various data types while requiring low computational overhead.


\subsection{Sensor-Aware Early Exit CNNs}

This section describes the proposed Sensor-Aware Early Exit Neural Network (SEE Neural Network) approach for CNNs. We start by discussing the base architecture, followed by detailed explanations of the early exit and late input blocks. Subsequently, we discuss the training and inference considerations for the SEE-based CNNs.
\subsubsection{Base CNN Architecture }

We consider a generic CNN architecture that is widely used in literature, as illustrated in top of Figure~\ref{fig:overview}.
The CNN architecture consists of multiple processing layers, including convolutional filters, max pooling, activation functions, and normalization layers. These are designed to extract relevant features from the input data. 
The feature extraction stages are followed by multiple fully connected layers and a softmax classification block to determine the logits of each class. 
Class with the highest logit value is provided as the final label. As noted in the previous section, the traditional CNN architectures use complete data in a segment, thus increasing energy cost. To address this issue, subsequent sections introduce strategies for energy savings through the implementation of early exit and late input blocks. 


\subsubsection{Early Exit Layers}
Early exit neural networks have been proposed in prior approaches~\cite{teerapittayanon2016branchynet,li2023predictive}.
The fundamental concept involves integrating additional classification layers in intermediate layers of the network. This allows for the generation of class predictions before processing the entire network.

For instance, after two convolutional layers, we may include a set of fully connected layers and a softmax block. These early exit blocks provide class predictions without needing to execute all layers of the CNN. If the confidence level of predictions made by the early exit block meets a predefined threshold, the neural network will terminate early, foregoing further processing. Confidence is a key parameter as the CNN is more likely to make errors in initial layers due to less information being processed at this stage. One significant advantage of early exit neural networks is their ability to bypass the computational overhead associated with the later CNN layers. 

We propose to include early exit layers consisting of one convolutional filter, max pooling, and ReLU activation layers. The ReLU layer is followed by two fully connected layers and a softmax classification block. 
We choose early exit blocks with a single convolutional filter to ensure low overhead at runtime.

\begin{figure}[t]
    \centering
    \includegraphics[width=0.83\linewidth]{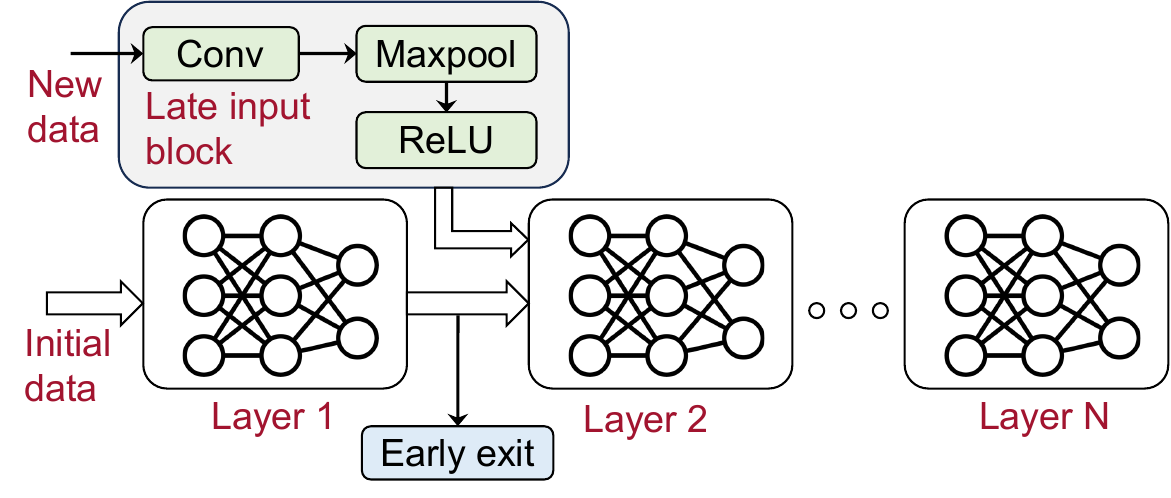}
    \vspace{-3mm}
    \caption{Illustration of the late input block}
    \label{fig:late_input}
\end{figure}

\subsubsection{Late Input Block}
A key component of the proposed SEE framework is the dynamic integration of input data across multiple layers as they become available. For example, 20\% of data is provided as input to the first layer, and an additional 20\% data is introduced at the third layer after the first early exit. The new data must be dynamically integrated. That is, the dimensions of new data values must match the data dimensions at later input layers. Moreover, the raw sensor data must be processed before being fed into the CNN layers. To this end, we propose late input blocks to feed partial segment data into the later layers.

Figure~\ref{fig:late_input} illustrates the main components of the late input blocks. Each late input block begins with a convolution filter to transform the input data into appropriate number of channels required for integration at an intermediate layer. Next, a max pooling layer is included to obtain appropriate data length for the intermediate layer. Finally, a ReLU activation layer is included to match the activation function of all other layers in the CNN. We note that the late input blocks may include multiple convolution and max pool layers. However, our experiments show that one layer is sufficient to maintain high accuracy without increasing the overhead.

The proposed SEE framework incorporates a late input block whenever new data are fed into the CNN after an early exit. These blocks correspond to layers immediately following the layers where early exits are implemented. This architecture is justified, as the need for new sensor data arises \textit{only} when the classifier cannot make a confident prediction with the available sensor data. We note that all remaining data in a segment are provided to the late input block at the last early exit. This ensures that the terminal exit in the CNN always has access to the complete data in a segment.


 

\subsubsection{SEE Neural Network Training}
CNNs are typically trained by optimizing the softmax cross-entropy loss. In particular, the optimization minimizes the cross-entropy between the true label and predicted label. While the cross-entropy optimization can be directly applied to CNNs with single exits, we must use additional loss functions to train the proposed SEE neural networks. Following the approach in~\cite{teerapittayanon2016branchynet} we can define the cross-entropy loss at any exit as:
\begin{equation}
    \mathcal{L}(\bm{y}, \bm{\hat{y}}_{\mathrm{exit}_n}, \theta) = - \frac{1}{|\mathcal{A}|} \sum_{i=1}^{|\mathcal{A}|} y_i log(\hat{y}_i)
\end{equation}
where $\bm{y}$ is one-hot encoded ground truth label, $\bm{\hat{y}}$ is the prediction, $\theta$ are the network weights, and $\mathcal{A}$ is the set of all labels. We can further write the prediction at an exit as:
\begin{equation}
    \bm{\hat{y}}_{\mathrm{exit}_n} = \mathrm{softmax}(\bm{z});\quad\mathrm{and}\quad \bm{z} = f_{\mathrm{exit}_n}(\mathbf{X_n}, \theta)
\end{equation}
where $f_{\mathrm{exit}_n}$ is the output of the $n^{th}$ exit and $\mathbf{X_n}$ is the input data vector for the $n^{th}$ exit.
That is, $\mathbf{X_n}$ denotes the partial input available for processing at the $n^{th}$ exit.

Our goal is to maximize the accuracy at all exits, we can combine the loss from each exit to form a weighted loss as:
\begin{equation}
    \mathcal{L}_{\mathrm{SEE}}(\bm{y}, \bm{\hat{y}}_{\mathrm{exit}_n}, \theta) = \sum_{n=1}^N \lambda_n \mathcal{L}(\bm{y}, \bm{\hat{y}}_{\mathrm{exit}_n}, \theta)
\end{equation}
where $N$ is the number of exits, including the terminal exit, and $\lambda_n$ denotes the weight assigned to $n^{th}$ exit.
The overall loss function is optimized during training to obtain the network parameters $\theta$. We first obtain the loss function at all exits during forward pass of the training.
That is, the network goes through all exits during training to capture the loss at each exit.
Then, the backward pass updates the parameters using gradient descent algorithms.
In this work, We use the Adam optimizer~\cite{kingma2014adam}, while noting that any CNN training algorithm is applicable.

\subsubsection{Sensor-Aware Inference}
The CNN model is optimized for efficient online inference using multiple exits and input blocks. The associated Algorithm~\ref{algo:runtime_see}, details the online inference steps for each new segment in an application. Initially, the algorithm feeds the first batch of partial data, denoted by $\mathbf{X}_1$ into the CNN.
Using this data, the CNN evaluates the class probabilities at the first exit and obtains a classification. If the confidence, determined by the entropy of the softmax outputs at this early exit, is sufficiently high,  the CNN will terminate processing  and deactivate the sensors until the next segment. This conditional exit process is outlined in  Line 9 of the algorithm.
 The entropy of the softmax outputs calculated the early exit is used as the confidence measure. We can write the entropy $E$ at an exit $n$ as:
\begin{equation}
    E(\bm{\hat{y}}_{\mathrm{exit}_n}) = -\sum_{i=1}^{|\mathcal{A}|} \hat{y}_i log(\hat{y}_i)
\end{equation}
High entropy indicates greater uncertainty in the predictions, with maximum entropy occurring when all class probabilities are equal Consequently, if the entropy falls below a predefined threshold, it indicates sufficient confidence for the classifier to make an early exit, conserving energy.

If the entropy exceeds this threshold, the algorithm supplies the next batch of partial data to the CNN. This iterative process of evaluating and potentially exiting continues until an early exit occurs or the final classification is determined at the terminal exit. A key advantage of the SEE model is its ability to make early exits, thereby significantly reducing sensor operation time and conserving energy.

\begin{algorithm}[t]
\small
	\caption{\small SEE Runtime Inference} \label{algo:runtime_see}
\SetAlgoLined
\textbf{Input:} Trained classifier $\theta$, Exit thresholds $\mathbf{T}$\\
\For {$n = 1, 2, \dots, N$}  {
$\mathbf{X_n} \gets$ Obtain partial data for current exit \\
$\bm{z} \gets f_{\mathrm{exit}_n}(\mathbf{X_n}, \theta)$\\
$\bm{\hat{y}}_{\mathrm{exit}_n} \gets \mathrm{softmax}(\bm{z})$ \\
 $E(\bm{\hat{y}}_{\mathrm{exit}_n}) \gets -\sum_{i=1}^{|\mathcal{A}|} \hat{y}_i log(\hat{y}_i)$ \\
\If {$E(\bm{\hat{y}}_{\mathrm{exit}_n}) < T_n$} {
Put sensors in low power mode \\
\textbf{return} Label prediction \\
}
}
\textbf{return} Label prediction \\
\end{algorithm}

\subsection{Sensor-Aware Random Forest}\label{sec:rf}
This section outlines the SEE approach for random forest classifiers, detailing the base architecture and the integration of early exits and late inputs.
\subsubsection{Base Random Forest Architecture}

We start with a baseline random forest architecture that takes all data in a segment to provide classifications. The random forest is chosen such that it provides high accuracy while having a minimal number of trees and levels in each tree. The baseline random forest architecture provides accuracy when using all data in a segment to evaluate the efficacy of early exits.


\subsubsection{Early Exit and Late Input for Random Forest}
Unlike neural network architectures, random forest classifiers do not provide the ability to insert early exits or late input blocks in intermediate levels of a tree. Therefore, we utilize an ensemble of random forests to enable early exit and late input for SEE.
Specifically, we obtain multiple random forest classifiers corresponding to each combination of partial data input. The classifiers successively increase in 
size to account for increasing amount of sensor data.
Moreover, we ensure that the total size of random forest ensemble is smaller than the base classifier to avoid additional memory overhead.
Each random forest classifier is trained to individually maximize the classification accuracy with partial sensor data. For instance, the first classifier may use 20\% of the data, while second classifier may use the first 20\% and additional 30\% of data.
Trained ensemble of the random forest classifier is deployed on the device to perform runtime classification.

\subsubsection{Runtime Random Forest Inference}
SEE uses a procedure similar to Algorithm~\ref{algo:runtime_see} to perform energy-efficient inference using random forests.
We start with the smallest classifier that uses the first batch of partial data.
If the entropy of class probabilities is less than a threshold, the algorithm exits and turns off sensors. If the classifier is not confident, next random forest in the ensemble is invoked.
The procedure continues until the classifier is confident or all data are processed at the last random forest.

\subsection{SEE Hyperparameter Optimization}
The SEE approach has the following hyperparameters that must be chosen carefully to maximize accuracy while minimizing overhead.

\noindent\textbf{Number of early exits and placement:} The number of early exits and their placement within a CNN or random forest must be chosen carefully to obtain high accuracy.
Placement of exits early in the network can lead to lower accuracy due to insufficient processing, while later exits may not provide sufficient reduction in energy. Similarly, large number of early exits increase overhead.

\noindent\textbf{Partial data input at each exit:} The amount of partial data from a segment input at each is a crucial parameter in the proposed SEE approach. Providing few data points in initial layers can hamper accuracy due to insufficient data, while large portion of segments in early exit layers reduce energy savings.

\noindent\textbf{Exit thresholds:} Entropy thresholds for early exits are another important parameter that affect accuracy and energy efficiency. We would like the early exits to provide high true positive rates and minimize false positives.

\noindent\textbf{Exit loss weights:} Loss weights $\lambda_n$ are required to train the early exit CNNs in the proposed framework. Appropriately assigning loss weights is important to ensure that all exits provide high accuracy.
Following recommendations in~\cite{teerapittayanon2016branchynet} we use decreasing weights from the first to last exit.

We choose the hyperparameters for SEE classifiers by performing design space exploration with a range for each parameter. Details on the design space exploration are provided in Section~\ref{sec:experiments}.


\section{Experimental Evaluation} \label{sec:experiments}

\subsection{Experimental Setup}\label{sec:exp_setup}
This section outlines  our experimental setup used to validate the SEE classifiers. It includes details on the device model, datasets, classifier representation, and evaluation metrics employed in our study. 

\subsubsection{Device Model}
For our experiments, we employed the Nvidia AGX Xavier~\cite{Xavier} board, a widely recognized platform tailored for edge applications. This device combines Nvidia graphics processors, suitable for ML models, with ARM processor cores. Specifically, it integrates eight Nvidia Carmel processors to enable sensor data processing. 
We utilize the Nvidia device to measure the overhead of proposed SEE classifiers while noting that any embedded processor is applicable.
The Nvidia AGX provides us with a common platform to evaluate SEE  across diverse datasets with various requirements.


\subsubsection{Datasets}
To validate efficacy of the models, we conduct experiments with six diverse activity and health datasets, considering the energy constraints inherent in low-power wearable devices used for health monitoring. The datasets used are the following:

   \vspace{1mm}
    \noindent \textbf{\textit{WESAD}}~\cite{schmidt2018introducing}: WESAD is a multi-modal dataset using wearable sensors for affect detection. Collected from 15 participants experiencing three different affective states (neutral, stress, amusement), it employs five different sensors placed on the chest: electrocardiogram (ECG), electrodermal activity (EDA), electromyogram (EMG), respiration (RESP), and body temperature (TEMP). 

    \vspace{1mm}
    \noindent \textbf{\textit{Physical Activity Recognition (Shoaib)}}~\cite{shoaib2014fusion}: This dataset is essential for healthcare tasks as understanding a patient's activities is crucial for managing movement disorders and rehabilitation [\textit{Maetzler et al., 2016}]. The Shoaib dataset provides accelerometer data for 10 users performing seven activities.

    \vspace{1mm}
    \noindent \textbf{\textit{PAMAP2}}~\cite{reiss2012introducing}: Another activity recognition dataset, PAMAP2, offers data from three accelerometers for five activities \{\textit{lying, sitting, walking, running and cycling}\} with nine users. The sensors are placed on the wrist of the dominant arm, chest, and the dominant side's ankle.

    \vspace{1mm}
    \noindent \textbf{\textit{EMG Physical Action (EMG)}}~\cite{misc_emg_physical_action_data_set_213}: The EMG dataset focuses on activity recognition for users who may exhibit aggression during tasks. It collects data using EMG sensors, monitoring activities such as Walking, Kicking, Jumping, and Headering. The dataset includes recordings from eight sensors placed on the upper arms and legs of the users.

    \vspace{1mm}
    \noindent \textbf{\textit{Epilepsy}}~\cite{villar2016generalized}: This dataset, collected from six participants using an accelerometer on the dominant wrist, covers four different activities: walking, running, sawing, and seizure mimicking.

    \vspace{1mm}
    \noindent \textbf{\textit{SelfRegulationSCP1 (SR-SCP1)}}~\cite{birbaumer2001brain}: Electroencephalography (EEG) is commonly used in brain-machine inference tasks. We employ the SR-SCP1 dataset, which includes EEG data from six channels used in a control system driving spelling devices for paralyzed patients.

\subsubsection{Base CNN and Random Forest Classifiers}
In the SEE classifier, we employ one-dimensional (1-D) CNNs as they are effective for time-series applications. They deliver promising results while requiring less computational power than 2-dimensional (2-D) CNNs~\cite{singh20211d,kiranyaz20211d}.  
Our specific implementation of baseline 1-D CNNs uses five convolutional, max pool, and ReLU layers followed by two fully connected layers. Classification is performed using a final softmax layer. The baseline CNN is adapted to incorporate early exit and late input layers.

The CNNs are implemented using the PyTorch library~\cite{paszke2019pytorch} in Python. The Adam optimizer~\cite{kingma2014adam} is used to train the CNNs for all datasets. For each dataset, 60\% of the data is used for training, while remaining 40\% is reserved for testing.

For the random forest implementation in the SEE framwork, we utilize the Scikit-learn library~\cite{scikit-learn}. As noted in Section~\ref{sec:rf}, an ensemble of random forests is trained to achieve early exit and late input. Similar to CNNs, 60\% of the data in each dataset are used for training, while remaining 40\% are used for testing.

\subsubsection{Evaluation Metrics} 
The primary evaluation metrics are classification accuracy and sensor energy consumption. The objective is to maximize the classification accuracy while reducing sensor energy consumption relative to classifiers that rely on entire segments of data. The secondary evaluation metric is the memory overhead of the proposed SEE classifiers.
\begin{table}[b]
\caption{Summary of range of hyper parameters used in SEE }
\label{tab:param}
\begin{tabular}{ll}
\hline
Parameter             & Range    \\ \hline
Data percentage       & 10 -- 50   \\
Number of early exits & 1 -- 2   \\
Confidence threshold  & 0.1 -- 1.5 \\
Loss weight           & 1 -- 4     \\ \hline
\end{tabular}
\end{table}

\subsection{SEE Hyperparameter Selection}
The first step in the SEE model training is choosing appropriate hyperparameters 
including the number of early exits, proportion of data in late input blocks, confidence thresholds, and loss weights.
Table~\ref{tab:param} shows the range of parameters explored for the design space exploration in SEE. 
We also explore different numbers of trees in the random forest classifier to form an ensemble that provides high accuracy with low sensor energy consumption.
We note that loss weights are applicable only for the CNNs since random forest classifiers are trained independently.

For CNNs, we limit design space exploration to at most two early exits as adding a  higher number of exits does not show improvements in accuracy or energy efficiency. In contrast, for random forest classifiers, we explore up to four early exits since the ensemble structure reduces overhead per classifier, making additional exits beneficial. 

The design space exploration starts by training 1-D CNNs and random forests for each combination of exit placements, proportion of data in late input blocks and loss weights.
The trained classifiers are then tested with each combination of confidence thresholds to obtain the classification accuracy and potential energy savings.
Finally, classifiers that provide the highest accuracy and energy savings are chosen for deployment in the application.

\begin{figure}[t]
    \centering
    \includegraphics[width=1\linewidth]{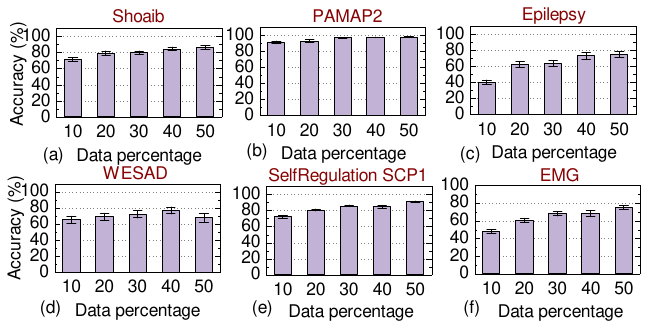}
    \caption{Accuracy Comparison with different input data percentage for the neural network for one early exit}
    \label{fig:percentage_2}
\end{figure}

\begin{figure}[t]
    \centering
    \includegraphics[width=1\linewidth]{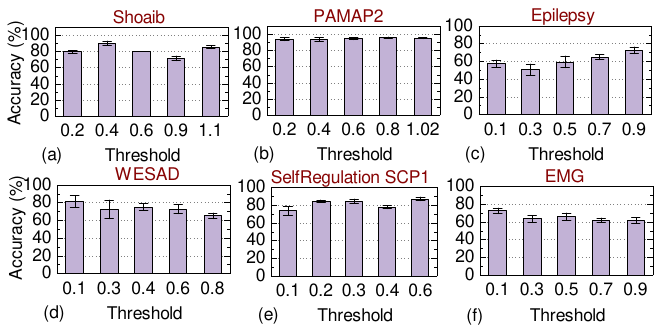}
    \caption{Accuracy Comparison with different threshold for the neural network for one early exit}
    \label{fig:threshold_2}
\end{figure}
\subsubsection{Sensitivity of Accuracy to Hyperparameters}
This section examines how the accuracy of 1D-CNNs with a single early exit is influenced by the proportion of input data and confidence thresholds.
Figure~\ref{fig:percentage_2} illustrates how accuracy changes as varying amounts of data are fed into the first CNN layer. As expected, the accuracy increases as more data is provided in these initial layers. Remarkably, using 30--40\% of the data yields an accuracy  80\% or higher, achieving up to 60--70\% in sensor energy savings.

Figure~\ref{fig:threshold_2}  examines how the mean accuracy is affected by varying confidence thresholds for early exits. Recall that lower threshold values necessitate higher confidence from the classifier before taking an early exit. For most datasets, lower thresholds correspond to higher classification accuracy, which aligns with expectations since the classifier exits \textit{only} when it is highly confident.

The Shoaib dataset shows higher variance in accuracy with changes in threshold values. This can be attributed to the fact that different training iterations may learn slightly different CNNs, thus leading to variance in their accuracy.
However, lower threshold values also decrease energy savings, as early exits become less frequent. Therefore, we aim to choose threshold values that balance accuracy in early exits and energy savings.

\subsection{Accuracy Analysis}
This section analyzes the accuracy for the classifiers chosen at the end of the design space exploration in the previous section.
We start with accuracy of the 1-D CNN in Table~\ref{tab:CNN_acc}. 
The chosen CNN architecture has two early exits at layers two and four, respectively for Shoaib and Epilepsy datasets. For SCP, we have the early exits at layers two and three. The rest of the datasets have early exits at layer three and four.
30 or 40\% of the data are provided at the beginning, while an additional 30 to 40\% data are provided after layer two or three.
Remaining data are provided in layer three or four.
The table shows that accuracy with SEE CNN is within 1\% of the baseline accuracy with all datasets.
The accuracy is even higher for the majority of the datasets.
The higher accuracy stems from the fact that the CNN is able to better train later layers of the CNN to classify difficult classes with high accuracy.
Overall, the 1-D CNN classifiers with SEE obtain accuracy that is comparable to the baseline classifiers.

Next, Table~\ref{tab:RF_acc} shows accuracy with the chosen random forest classifier.
We obtain the best accuracy and energy savings trade-off with a classifier that provides 20\% data for the first random forest in the majority of cases.
We see that the accuracy is comparable to the baseline for all datasets, except the SelfRegulation SCP1 dataset. The lower accuracy is due to the fact that the earlier random forests make some inaccurate predictions with early exits.
The accuracy is still within 5\% of the baseline random forest with 100\% segment data.
In summary, the proposed SEE classifiers provide accuracy comparable to baseline classifiers with partial data.

\begin{table}[]
\caption{Accuracy comparison between baseline and SEE CNN classifier}
\label{tab:CNN_acc}
\begin{tabular}{lrr}
\toprule
                         & \multicolumn{2}{c}{Model accuracy (\%)}                                      \\ \midrule
Datatset            & Baseline & SEE CNN\\ \midrule
Shoaib              &    96           &   98              \\ 
PAMAP2              &      98         &     99            \\ 
Epilepsy            &      94         &     99            \\ 
WESAD               &      98         &     99            \\ 
SelfRegulation SCP1 &       90         &    96             \\ 
EMG                 &       94        &      91           \\ \bottomrule
\end{tabular}
\end{table}

\begin{table}[]
\caption{Accuracy comparison between baseline and SEE RF classifier}
\label{tab:RF_acc}
\begin{tabular}{lrr}
\toprule
                         & \multicolumn{2}{c}{Model accuracy (\%)}                                      \\ \midrule
Datatset            & Baseline & SEE RF\\ \midrule
Shoaib              & 97              & 96                \\ 
PAMAP2              & 98              & 98                \\ 
Epilepsy            & 85              & 91                \\ 
WESAD               & 99              & 99                \\ 
SelfRegulation SCP1 & 85               & 80                \\ 
EMG                 & 90              & 89                \\ \bottomrule
\end{tabular}
\end{table}

\begin{figure}[t]
    \centering
    \includegraphics[width=0.99\linewidth]{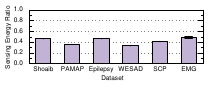}
    \vspace{-5mm}
    \caption{Energy savings ratio in CNN classifier}
    \label{fig:NN_energy}
\end{figure}

\begin{figure}[t]
    \centering
    \includegraphics[width=0.99\linewidth]{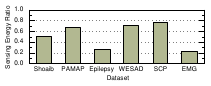}
    \vspace{-5mm}
    \caption{Energy Savings Ratio in random forest classifier}
    \label{fig:RF_energy}
\end{figure}

\subsection{Energy Savings from SEE Classifiers}
One of the key contributions of the proposed approach is the potential of sensor energy savings by using partial data and early exits. This section delves  into the analysis of the energy savings achieved by the proposed SEE classifiers.
Figures~\ref{fig:NN_energy} and~\ref{fig:RF_energy} depict the ratio of sensor energy used by SEE in comparison to the baseline classifier.
Specifically, these figures illustrate the ratio of sensor energy consumed in SEE versus the baseline. Lower numbers indicate that SEE consumes less energy for sensing.
The figures encompass all classifier combinations that exhibit accuracy within 1\% of the baseline with 100\% segment data.

It is evident that both SEE classifiers exhibit significantly lower energy consumption when compared to the baseline. Indeed, the energy ratio is lower than 0.5 for all datasets using the 1-D CNN.
This means that we can achieve more than 50\% savings in sensing energy.
Similar results are observed for the random forest classifiers. Some datasets exhibit lower energy savings due to the more frequent use of later exits. 
Consequently, this leads to diminished savings in energy consumption.
In summary, these results show that the proposed SEE classifiers enable significant energy savings for time-series applications.

\subsection{Overhead of SEE Classifiers}
Classifiers for health applications are typically deployed on resource-constrained devices. 
Therefore, we analyze the memory overhead of SEE networks in this section.
Memory overhead in SEE CNNs consist of the weights required for the early exit and late input blocks.
Similarly, the memory overhead for random forest classifiers consists of additional classifiers required to construct the ensemble.
Table~\ref{tab:overhead} shows the memory overhead for all datasets.
We see that the CNNs have higher memory requirements compared to the baseline classifiers. However, the overhead is within the memory available in the Nvidia Xavier devices.
The overhead of random forest classifiers is significantly lower due to lower memory requirements of random forests.

Energy overhead of the proposed networks is also negligible since they add minimal computations compared to the default classifiers. Indeed, our evaluations show that the energy consumption of additional blocks is negligible compared to the energy savings enabled through partial data processing.
In summary, the proposed approach enables significant energy savings for sensing without impacting the accuracy values.

\begin{table}[t]
\caption{Summary of memory overhead of baseline and SEE classifiers }
\label{tab:overhead}
\begin{tabular}{lrrrr}
\hline
                    & \multicolumn{4}{c}{Memory Storage (KB)}            \\ \hline
Datatset            & Baseline CNN & SEE CNN & Baseline RF & SEE RF \\ \hline
Shoaib              & 723         & 1863    &597 &  1099      \\
PAMAP2              & 1706         & 4498    &  190           & 389       \\
Epilepsy            & 329         & 712    &   89          &  208      \\
WESAD               & 329         & 714    &    0.18
         &  30      \\
SCP & 1050         & 2673    &    9         &   17     \\
EMG                 & 330        & 547    &    96         &   341     \\ \hline
\end{tabular}
\end{table}

\section{Conclusions and Future Work} \label{sec:conclusion}
Several real-world applications require time-series processing for identification of relevant parameters.
ML models used for time-series applications typically use data from complete segments for classification.
This can lead to high sensing energy consumption for the device. 
We proposed sensor-aware early exit classifiers to utilize partial data in making classification decisions. 
The key idea in the proposed approach is to utilize partial data in a segment to perform classification for easier classes.
Experimental results with six diverse datasets showed that the proposed approach enables up to 50-60\% energy savings on average for sensing.
Our immediate future work is to extend the approach to multiple classifiers and deploy them in real-world settings.


\footnotesize{\bibliographystyle{abbrv}

\bibliography{references/IEEEabrv,references/health_refs,references/embedded_refs,references/flexible,references/Sensor_aware,references/wearable_iot}

\end{document}